%% file: eccv2012_83_v5.tex
\pdfoutput=1
\documentclass[runningheads]{llncs_footnotesize}
\usepackage{graphicx}
\usepackage{amsmath,amssymb} 
\usepackage{color}
\usepackage{array}
\usepackage{addcmds}
\usepackage[width=122mm,left=12mm,paperwidth=146mm,height=193mm,top=12mm,paperheight=217mm]{geometry}

  \renewenvironment{thebibliography}[1]{%
    \begin{oldthebibliography}{#1}%
      \setlength{\parskip}{1pt}%
      \setlength{\itemsep}{0ex}%
  }%
  {%
    \end{oldthebibliography}%
  }

\begin{document}

\newcommand{\pmm}[1]{\ensuremath{\boldsymbol{#1}}} 
\newcommand{\pv}[1]{\ensuremath{\boldsymbol{#1}}} 
\newcommand{\ps}[1]{\ensuremath{\mathcal{#1}}} 
\newcommand{\transpose}[0]{\ensuremath{\mathsf{T}}} 

\newcommand{\figspcYcaption}{\vspace*{-10pt}}
\newcommand{\figspcYabove}{\vspace*{-12pt}}
\newcommand{\figspcYbelow}{\vspace*{-15pt}}
\newcommand{\secspcYabove}{\vspace*{-11pt}}
\newcommand{\secspcYbelow}{\vspace*{-8pt}}
\newcommand{\subsecspcYabove}{\vspace*{-8pt}}
\newcommand{\subsecspcYbelow}{\vspace*{-4pt}}
\newcommand{\subsubsecspcY}{\vspace*{-13pt}}
\newcommand{\eqnspcYabove}{\vspace*{-7pt}}
\newcommand{\eqnspcYbelow}{\vspace*{2pt}}
\newcommand{\parspcY}{\vspace*{-5pt}}

\pagestyle{headings}
\mainmatter
\def\ECCV12SubNumber{83}  

\title{Detection of Partially Visible Objects} 


\author{Patrick Ott\inst{1}\and Mark Everingham\inst{1}\and Jiri Matas\inst{2}}

\institute{School of Computing, University of Leeds\\
\and
Center of Machine Perception, Czech Technical University, Prague\\}

\maketitle

\vspace{-20pt}
\begin{abstract}
An `elephant in the room' for most current object detection and
localization methods is the lack of explicit modelling of partial
visibility due to occlusion by other objects or truncation by the
image boundary. Based on a sliding window approach, we propose a
detection method which explicitly models partial visibility by
treating it as a latent variable. A novel non-maximum suppression
scheme is proposed which takes into account the inferred partial
visibility of objects while providing a globally optimal solution.
The method gives more detailed scene interpretations than
conventional detectors in that we are able to identify the
visible parts of an object. We report improved average precision
on the PASCAL VOC 2010 dataset compared to a baseline detector.
\end{abstract}

\vspace{-15pt} \secspcYabove\section{Introduction}\secspcYbelow
\label{sec:intro}

One aspect of object appearance which can greatly affect the
success of any object detector, yet which has largely been
overlooked in existing methods, is \emph{partial visibility}.
State-of-the-art methods~\cite{Felzenszwalb10,Vedaldi09} generally
assume that the entire object instance is visible or at
most offer a way to model truncation of an object by the image
boundary~\cite{voc-release4,Vedaldi09b}. Moreover these approaches
predict a bounding box for a partially visible object which
`hallucinates' the hidden parts of the object.

\begin{figure}[b]
    \figspcYabove
        \centering
        \begin{tabular}{ccccc}
            \includegraphics[scale=0.225]{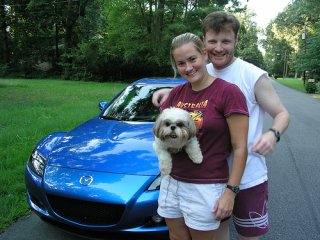}
            &
             \includegraphics[scale=0.225]{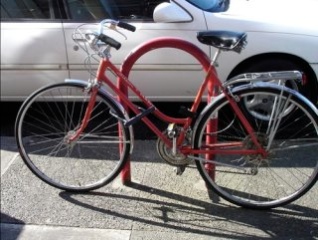}
            &
            \includegraphics[scale=0.225]{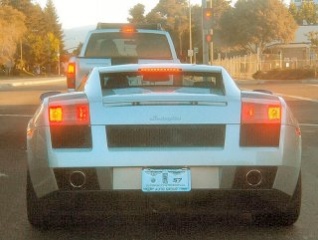}
            &
            \includegraphics[scale=0.225]{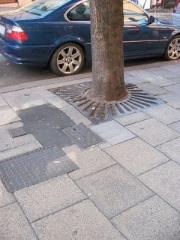}
        &
            \includegraphics[scale=0.225]{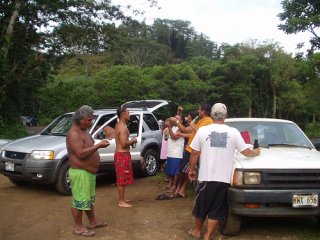}
            \\
            (a) & (b) & (c) & (d) & (e)
        \end{tabular}
    \figspcYcaption
        \caption[Examples of partial visibility of the car
        class]{Partial visibility in natural images is commonplace
        \eg cars can be occluded by other cars (c) or other objects
        (a, b, d, e), or truncated by the image boundary (b).}
        \label{fig:ExampleOcclusions}
    \figspcYbelow
\end{figure}

We propose a model for detection which explicitly accounts for
partially visible objects. Partial visibility of objects in a
scene is commonplace, for example as shown in
Fig.~\ref{fig:ExampleOcclusions} cars are routinely occluded by
other objects such as other cars~(c), people~(a, e) or trees~(d),
or may lie partially outside the image~(b, d).

Current object detection methods typically neglect the possibility
of partial visibility of an object, hoping that the feature
representation of a partially visible object is sufficiently
unaffected that the object/non-object classifier function will
still exceed the threshold \ie\ the detector still `fires'. While this
is a reasonable assumption for small occlusions, for example a
tree in front of a car (Fig~\ref{fig:ExampleOcclusions}~(d)), it
is clearly unrealistic when a larger part of the object is hidden,
see \eg\ (a, c). In practice, trained object detectors usually
place high weight on a few key regions of the object that are
characteristic of the object category, \eg wheels for a car. When
these regions are occluded (\eg
Fig.~\ref{fig:ExampleOcclusions}~(a)), the corresponding positive
contribution to the detection score is lost, and given sufficient
missing regions, the image is misclassified as `non-object'. The
problem is compounded by the use of discriminative classifiers,
since the occluders will themselves typically contribute
\emph{negative} evidence to the presence of the object to be
detected.

\subsubsecspcY\subsubsection{Contributions.}

We make three key contributions: (i)~we propose a detection method
which explicitly accounts for partial visibility at both training
and test time, by modelling the visibility as a \emph{latent
variable}; (ii)~the proposed method gives a more accurate
interpretation of the imaged scene by not only reporting the
localization of an object but also by inferring which parts of the
object are visible and which are not; (iii)~We show that the
modelling of partial visibility results in improved performance in
 object detection.

\begin{figure}[t!]
    \figspcYabove
        \centering
     \includegraphics[scale=0.325]{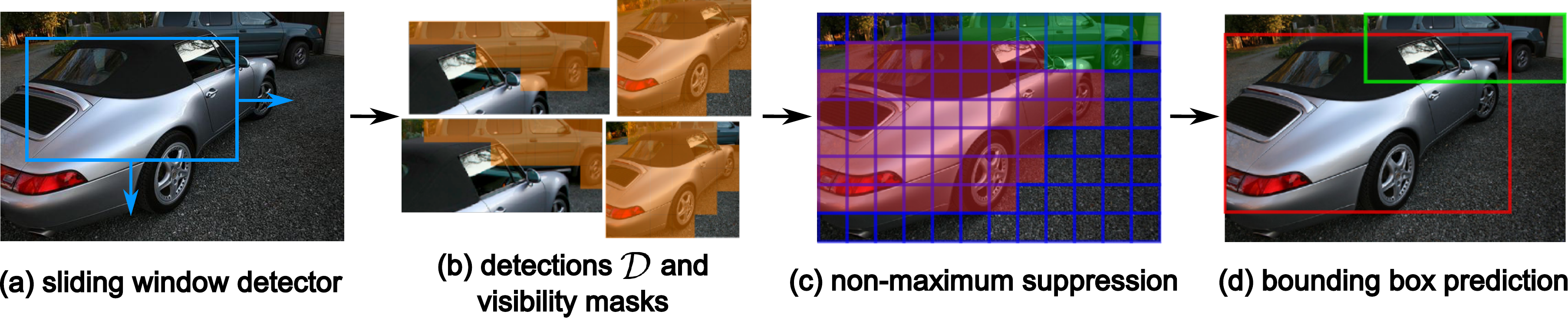}
    \figspcYcaption
        \caption[Pipeline of proposed detection framework]{The pipeline of the proposed detection framework: A sliding-window detector~(a) generates candidate detections~(b) each comprising a detection score and a block-wise mask of visibility inferred using an MRF model. Non-maximum suppression uses the additional cues of partial visibility for each window to infer a scene interpretation~(c), from which accurate object bounding boxes~(d) are predicted.}
        \label{fig:Overview}
    \figspcYbelow
\end{figure}

Fig.~\ref{fig:Overview} provides an overview of the proposed
framework. The method is based on a discriminative sliding window
approach (Fig.~\ref{fig:Overview}~(a)), taking advantage of an
holistic description of the object, but \emph{explicitly}
modelling partial visibility. For a given window we treat partial
visibility as a latent variable which specifies which regions of
the hypothesized object are visible or hidden. A prior over
plausible values of the variables is defined in the form of an
homogeneous Markov Random Field (MRF), capturing two properties:
(i)~the effect of partial visibility of an object class on the
classification score if a region of the window, \eg\ a car wheel,
is considered occluded; (ii)~spatial continuity of partial
visibility patterns within a window -- neighboring regions are
often occluded together.

At test time we infer partial visibility while simultaneously
computing a score representing the confidence that the object
class is present -- Fig.~\ref{fig:Overview}~(b) shows some example
detections with inferred visibility. Similar to other object
detectors~\cite{Dalal05,Felzenszwalb10}, the proposed method gives
multiple detections at the true location of an object instance.
Consequently, we propose a non-maximum suppression (NMS) scheme
which makes use of the pattern of visibility inferred for each
window -- see Fig.~\ref{fig:Overview}~(c). In this way our
approach can predict the correct visible extent of an object
taking into account hidden parts, and correctly resolve detections
of neighboring objects which may be discarded by conventional NMS
schemes which assume objects to have fixed `average' extent -- see
Fig.~\ref{fig:Overview}~(d).

\subsubsecspcY\subsubsection{Related Work.}

Rather little work has explicitly considered the problem of
partial visibility in general object detection.
Wang~\etal~\cite{Wang09} train global pedestrian detectors and
compute a patch-wise response map for by `splitting' the linear
classifier coefficients into patches. The mean shift algorithm is
then applied to infer un-occluded regions, for which individual
part-detectors are invoked. Fergus~\etal~\cite{Fergus03} propose a
generative model for object detection based on a constellation of
parts. They define a joint probability density over shape,
appearance, scale and part-level occlusion. The method uses a
probability table for all possible patterns of occlusion, which is
treated as a model parameter. Winn~\etal proposed the `layout
consistent random field' model~\cite{Winn06} which combines local
part detectors with a CRF of part configuration, placing
constraints on neighboring parts. The method offers an elegant
formulation for patch-wise scene labelling accounting for partial
visibility in the binary potentials of the CRF. The method relies
on somewhat weak part detectors, is limited to single-scale
objects, and inference in the model is computationally expensive.

Recently Gao~\etal~\cite{Gao11} presented a method closely
related to ours in that it treats visibility as a latent variable.
Different to our approach, the method requires training data to
have additional annotation of visible regions, where we infer this
information. Gao~\etal use a structured learning framework and
employ a loss function which is based on the overlap of a
predicted detection with a ground-truth object. This
overlap measure does not take into account the partial visibility
of an object, which we exploit in our proposed NMS scheme.

In the related area of face detection a number of authors have
investigated models of occlusion, but the applicability of these
approaches to general object detection has not been established.
Williams~\etal~\cite{Williams04} use a variational Ising
classifier to model contamination (\eg occlusion) by a binary
mask, and extend a kernel classifier for `clean' data to one that
can tolerate contamination. In contrast to our method, which
considers occlusion at training and test time, their method
functions as an extension to an occlusion-unaware face detector.
Lin \& Fuh~\cite{Lin04} adapt a boosted cascade to detect occluded
faces by `hard-coding' eight different types of partial visibility
and testing for them as well as for fully visible faces. The
method does not generalize beyond the hard-coded occlusion types.
In contrast, our method learns the patterns of occlusion.

\subsubsecspcY\subsubsection{Motivation.} 
\begin{figure}[t]
    \figspcYabove
        \centering
\begin{tabular}{cc}
            \includegraphics[scale=0.35]{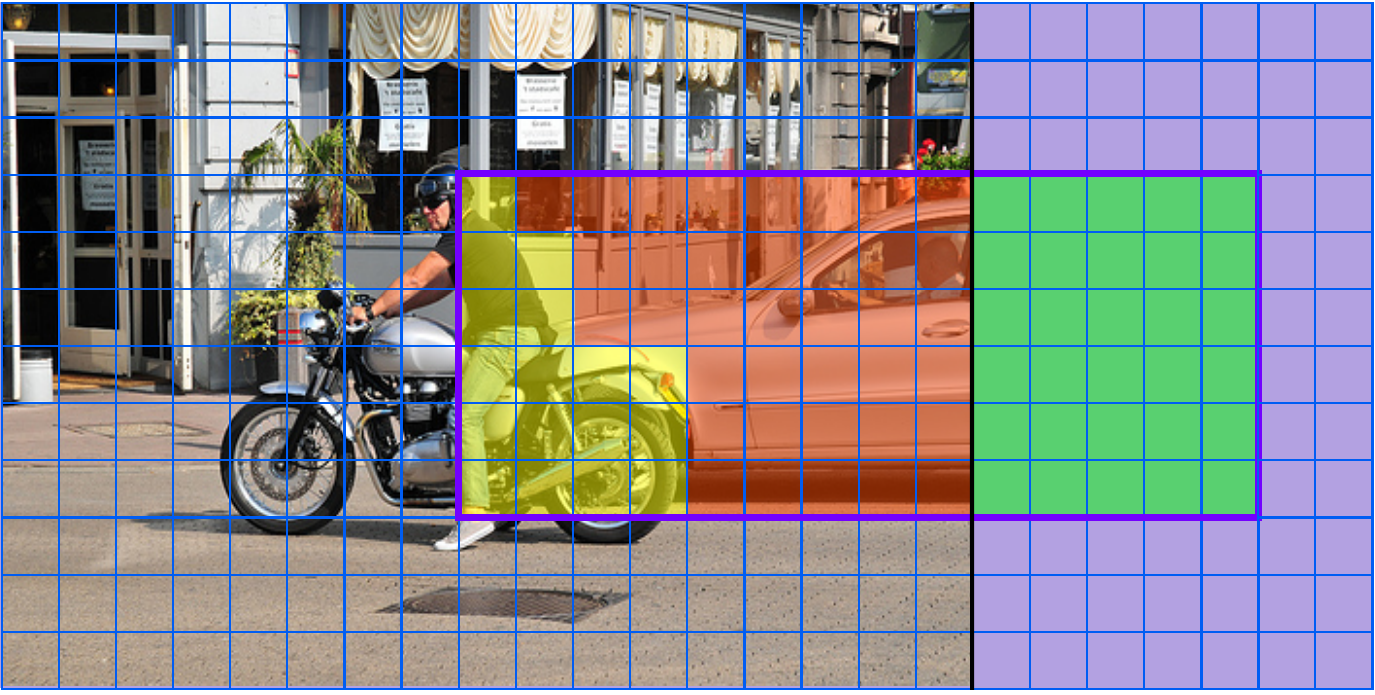} \hspace{5pt}
            &
    \includegraphics[scale=0.9]{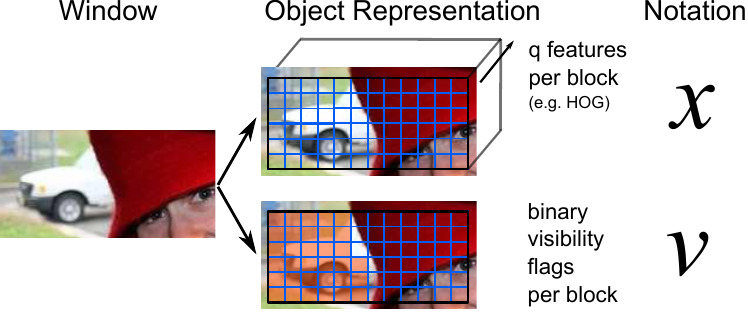}
    \\
    (a)~An example detection
    &
    (b)~Modelling of partial visibility
\end{tabular}
\figspcYcaption \caption{(a)~Our approach is motivated by the
method of Vedaldi~\&~Zisserman~\cite{Vedaldi09b} which pads the
image representation with additional blocks (purple region) to
detect truncated objects. During classification
Vedaldi~\&~Zisserman~\cite{Vedaldi09b} `count' the number of
patches outside of an image (green) and apply a learned classifier
weight to this number of patches. We adopt this approach but also
infer the visible parts of an object instance inside the image
(orange region) as well occluded parts of the object instance
(yellow). (b)~A window is represented by a block structure,
consisting of (i)~a $q$-dimensional feature descriptor per block,
concatenated into the feature descriptor for the entire window
\pv{x}; (ii)~a vector $\pv{v}$ of binary visibility flags, with
one flag per block. The inferred visibility flags -- visible
blocks are shaded orange -- control which of the features are used
by the classifier, and are subject to a prior defining plausible
patterns of visibility. } \label{fig:methodoverview} \figspcYbelow
\end{figure}

Our approach is related
to the work of Vedaldi~\&~Zisserman~\cite{Vedaldi09b} and
Girshick~\etal~\cite{voc-release4}. Both methods propose
part-based models for object detection and pad the image
representation with additional blocks to detect truncated objects.
This is illustrated in Fig.~\ref{fig:methodoverview}~(a) -- purple
blocks indicate patches outside the image.
Vedaldi~\&~Zisserman~\cite{Vedaldi09b} `count' the number of
blocks outside an image which are covered by the window (green in
Fig.~\ref{fig:methodoverview}~(a)) and learn a classifier weight
for this number of patches which compensates for the missing block
responses. Girshick~\etal~\cite{voc-release4} learn individual
bias terms for each block outside the image which are added to the
window score to compensate for the missing block responses. Both
methods give comparable results for the detectors used in this
paper. We adopt the approach of
Vedaldi~\&~Zisserman~\cite{Vedaldi09b} as it requires the learning
of fewer parameters. In addition to modelling truncation our
method also models partial visibility inside the image by treating
it as a latent variable (yellow/orange blocks in
Fig.~\ref{fig:methodoverview}~(a)).

\subsubsecspcY\subsubsection{Outline.}
We describe the proposed methods in Sections.~\ref{sec:method}, \ref{sec:learning} and \ref{sec:nms}. Sec.~\ref{sec:results} presents qualitative and quantitative results on the PASCAL VOC 2010 dataset~\cite{pascal-voc-2010}. Conclusions are offered in Sec.~\ref{sec:conclusions}.

\secspcYabove\section{Modelling Partial Visibility}\secspcYbelow
\label{sec:method}

Sliding window detectors generally move a window of fixed size
over an image (Fig.~\ref{fig:Overview}~(a)). For each window, a
feature vector is extracted and a classification function is
applied to classify the window as object/non-object. As shown in
Fig.~\ref{fig:methodoverview}~(b) we use features which can be
organized in a structure of blocks. We model partial visibility at
the level of blocks, \ie inference in our model yields a
per-window labelling of which blocks of the object are visible or
hidden. Throughout this paper the features of the $i$th block are
denoted $\pv{B}^i$ and the features of the entire window $\pv{x} =
\left\langle {\pv{B}^1 ,...,\pv{B}^r } \right\rangle$ for $r$
blocks.

\subsecspcYabove\subsection{Partial Visibility as a Latent Variable}\subsecspcYbelow
To represent partial visibility we append visibility flags $\pv{v}$ to the conventional feature representation $\pv{x}$ -- see Fig.~\ref{fig:methodoverview}~(b). If $v_i$ is switched off, \ie $v_i=0$, we consider the $i$th block to be not visible (occluded), if switched on ($v_i=1$) we consider it visible. The key to our method is to treat these visibility flags $\pv{v}$ as a latent variable to be inferred during classification of a window, allowing the classifier to operate on only the visible portion of the object, and enabling subsequent NMS processing to predict accurate bounding boxes for partially-visible objects. Furthermore, we additionally apply inference to \emph{training} examples to account for partially visible examples in the training data.

\subsubsecspcY\subsubsection{Unary Bias.} As noted in
Sec.~\ref{sec:intro} the features of an occluded block typically
contribute negatively to the classification score of a window.
Introducing visibility flags allows our method to correct this by
inferring which blocks are in fact not visible and `disabling' the
part of the classifier operating on the corresponding features.
This is achieved by replacing the partial classifier score for the
block with a \emph{learned} unary bias $u$, which compensates for
the missing image evidence~\cite{Vedaldi09b} in the case that a
block is inferred to be not visible ($v_i=0$).

\subsubsecspcY\subsubsection{MRF Prior.}
Partial visibility of natural objects tends to be contiguous, resulting in a large and connected area of invisible blocks, whereas complex or sparse patterns of partial visibility are not common. To model this notion of contiguity we define an MRF prior over $\pv{v}$ in the form of an Ising model~\cite{MacCormick98} which discourages assigning differing visible/invisible labels to neighboring blocks.

\subsecspcYabove\subsection{Window Responses}\subsecspcYbelow
Given a block representation of a window with features $\pv{x}$, we define the task of assigning a classification (object present) confidence to the window as maximization of a \emph{window response} function $g(\pv{x})$ \wrt the visibility flags $\pv{v}$ which define which blocks in the window are considered visible/hidden. The response function comprises (i)~a classification function and (ii)~a prior over visibility flags:
\begin{equation}
\eqnspcYabove
g\left( \pv{x} \right) = \mathop {\max }\limits_{\pv{v}} \left[ {f\left( {\pv{x},\pv{v}} \right) - \alpha \sum\limits_{\left( {i,j} \right) \in {\ps{N}}} {\rho \left( {{v_i},{v_j}} \right)} } \right]
\label{eqn:InferOcc}
\eqnspcYbelow
\end{equation}
where the first term $f(\pv{x},\pv{v})$ defines a `classification score' for the window \emph{given} the inferred pattern of visibility $\pv{v}$. We adopt a linear classification function, and later demonstrate the extension to a mixture of linear classifiers~\cite{Felzenszwalb10}:

\begin{equation}
\eqnspcYabove
f\left( {\pv{x},\pv{v}} \right) = b +  \sum\limits_{i = 1}^r {v_i \pv{w}^i  \cdot \pv{B}^i  + \left( {1 - v_i } \right)u}
\label{eqn:classfunc}
\eqnspcYbelow
\end{equation}

The visibility flags $v_i$ act as a selector/switch function -- if the visibility flag for the $i$th block $v_i$ is switched on, ${\pv{w}^i  \cdot \pv{B}^i }$ is added to the score, where $\pv{B}^i$ represents the features of the $i$th block and $\pv{w}^i$ the learned appearance of that block. If $v_i$ is switched off  the \emph{learned} unary bias $u$ is added to the score. $b$ is the bias term of the linear classification function.

The second term of the window response (Eqn.~\ref{eqn:InferOcc})
${\rho\left( {v_i,v_j }\right)}$ acts as a penalty by defining an
Ising prior~\cite{MacCormick98} over the field of visibility
flags. A penalty of $1$ is imposed if $v_i \ne v_j$ and $0$
otherwise for all pairs of neighboring blocks $\ps{N}$. We use a
conventional 4-neighborhood. Variable $\alpha$ defines the
relative weight of the classification score and contiguity terms.
If $\alpha$ is small, sparse patterns of visibility flags are
allowed, while large values of $\alpha$ encourage contiguous
patterns of occlusion. As noted, this model of visibility can be
interpreted as an MRF at the level of a single window, with nodes
corresponding to blocks.

\subsecspcYabove\subsection{Inference}\subsecspcYbelow
\label{sec:inference} The maximization in Eqn.~\ref{eqn:InferOcc}
can be solved efficiently using a graph-cut
algorithm~\cite{Boykov01b}, since the problem corresponds to a
standard binary-valued MRF, with sub-modular energy for positive
$\alpha$~\cite{Kolmogorov04}. However, despite the efficiency of
this method, an image comprises a large number of windows
($>10.000$) such that solving a graph-cut problem for each window
is still somewhat onerous. We therefore accelerate detection by
filtering each window using the following upper bound on $g\left(
\pv{x}\right)$:
\begin{equation}
\eqnspcYabove
\hat g\left( \pv{x} \right) = \mathop {\max }\limits_{\pv{v}} f\left( {\pv{x},\pv{v}} \right) \ge g\left( \pv{x} \right)
\label{eqn:upperbound}
\eqnspcYbelow
\end{equation}

This upper bound is derived from Eqn.~\ref{eqn:InferOcc} by
removing the binary terms since these are always nonnegative.
Maximization of the upper bound is achieved in a block-wise
fashion \ie the values for $v_i$ can be computed independently for
each block by comparing the classifier term for the block to the
bias term. If $\hat g\left( \pv{x} \right)$ scores above a
pre-defined threshold $t$, \eg $t=-1$, we compute $g\left( \pv{x}
\right)$.

\subsecspcYabove\subsection{Mixture Model}\subsecspcYbelow

Recent datasets for object localization, such as the PASCAL VOC
datasets~\cite{voc_ijcv}, include objects with substantial
variation in intra-class appearance due to both different types of
objects, \eg models of car, and large variations in viewpoint. To
cope with this variability we model a class as a `mixture model'
consisting of a set of linear classifiers (`mixture
components')~\cite{Felzenszwalb10}, where each component may be
specialized to a particular sub-class or viewpoint, learnt at
training time. The final classification confidence $h(\pv{x})$ for
a window is then computed as the maximum over all the component
classifiers $g^{i}(\pv{x})$:

\begin{equation}
\eqnspcYabove
h\left( \pv{x} \right) = \max _{i = 1 \ldots d} g^i \left( \pv{x}
\right)
\eqnspcYbelow
\label{eqn:mixmod}
\eqnspcYbelow
\end{equation}

In the following we refer to function $h\left(\pv{x}\right)$ as
the \emph{detector} and to the functions $g^{i}(\pv{x})$ as the
mixture components. The particular mixture component $m^*$
satisfying Eqn.~\ref{eqn:mixmod}, \ie ${m^*} = \mathop {\arg \max
}_m \left\{ {{g^{ m}}\left( \pv{x} \right)} \right\}$, is referred
to as the \emph{mixture assignment} of the example with features
$\pv{x}$. We use individual unary bias terms for each mixture
component because different mixture components might base their
classification decision on a different set of key regions,
requiring individual values of $u$ (Eqn.~\ref{eqn:classfunc}) to
compensate for the occlusion of such regions.

\secspcYabove\section{Learning}\secspcYbelow
\label{sec:learning}

In this section we cast the learning of detectors as an energy
minimization problem. We first introduce the necessary notation
and define an energy function for model training in
Sec.~\ref{sec:nrg} while optimization is discussed in
Sec.~\ref{sec:optim}.

\subsubsecspcY\subsubsection{Notation.}
A training dataset $\ps{T} = \left\{ {\left( {\pv{x}^1 ,y^1 } \right),...,\left( {\pv{x}^n ,y^n } \right)} \right\}$ is given where $y^k  \in \left\{ { + 1, - 1} \right\}$ is the class of the window (object/non-object) and $\pv{x}^k$ is the feature vector of the $k$th training example.  We represent the latent mixture assignments by $\ps{M} = \left\{ {m_1 ,...,m_n } \right\}$ and the visibility flags by $\ps{V} = \left\{ {\pv{v}^1 ,...,\pv{v}^n } \right\}$. They are combined into the set of latent variables $\ps{L} = \left\{ {\ps{M},\ps{V}} \right\}$. We recall that the mixture components of $h\left( \cdot \right)$ are essentially defined by the weight vector $\pv{w}$ and the unary bias term $u$. We represent these by $\ps{W} = \left\{ {\pv{w}^1 ,...,\pv{w}^d } \right\}$ and $\ps{U} = \left\{ {u^1 ,...,u^d } \right\}$ respectively. They are combined into the set of model variables $\ps{O} = \left\{ {\ps{W},\ps{U}} \right\}$.

We now present a learning scheme for estimating these model variables $\ps{O}$, while also inferring the latent variables $\ps{L}$ for each training example.

\subsecspcYabove\subsection{Energy Function}\subsecspcYbelow
\label{sec:nrg}
The energy function $E\left( {\left. \ps{O} \right|\ps{T}} \right)$ is dependent on the model variables $\ps{O} = \left\{ {\ps{W},\ps{U} } \right\}$ and assumes a training dataset $\ps{T}$ as given. It consists of two parts: (i)~a regularization term, ensuring good generalization on unseen data and (ii)~a loss term, determining how well the detector predicts the training labels:
\begin{equation}
\eqnspcYabove
E\left( {\left. \ps{O} \right|\ps{T}} \right) = \frac{\lambda }{2}\sum\limits_{j = 1}^d {\left[ {\left\| {\pv{w}^j } \right\|^2  + c_j \left( {u^j } \right)^2 } \right]}  + \sum\limits_{k = 1}^n {L\left( {y^k ,h\left( {\pv{x}^k } \right)} \right)}
\label{eqn:occnrg}
\eqnspcYbelow
\end{equation}
where $\left\| \cdot  \right\|$ is the $\ell_2$-norm. $\lambda$
sets the relative importance of the regularization term to the
loss term. We found only modest performance increases by
introducing separate regularization for $\pv{w}^j$ and $u^j$.
$c_j$ is the number of blocks in the $j$th mixture component; the
multiplication with $u^j$ ensures that the unary bias of each
mixture component is appropriately weighted in the regularization
term. The loss function $L\left( { \cdot } \right)$ determines how
deviations of the predictions of $h\left( \pv{x}^k \right)$ from
the true target value $y^k$ should be penalized -- we use the
hinge loss.

\subsecspcYabove\subsection{Optimization Scheme}\subsecspcYbelow
\label{sec:optim}

Optimizing $E\left( \cdot \right)$ \wrt the model variables
$\ps{O}$ while simultaneously inferring latent variables $\ps{L}$
for all training examples is a non-convex and discontinuous
optimization problem. However, we observe that when provided with
latent variables $\ps{L}$ for each training example in $\ps{T}$
the optimization problem takes a convex and continuous form. To
represent that latent variables $\ps{L}$ are fixed we extend the
notation of the energy function to $E\left( {\left. \ps{O}
\right|\ps{T},\ps{L}} \right)$. Given that observation we
implement an optimization scheme which alternates between
optimizing $E\left( {\left. \ps{O} \right|\ps{T},\ps{L}} \right)$
\wrt $O$ and updating the latent variables $\ps{L}$.

\subsubsecspcY\subsubsection{Classifier parameter updates and bootstrapping.}

For a fixed set of latent variables, we perform updates on
$\ps{O}$ by minimizing $E\left( {\left. \ps{O}
\right|\ps{T},\ps{L}} \right)$  using L-BFGS~\cite{Byrd95}. Having
trained a detector in this manner we extract high-ranking false
positives from windows not containing the target object class. We
add those false positives to the original training dataset
$\ps{T}$ and re-optimize $E\left( {\left. \ps{O}
\right|\ps{T},\ps{L}} \right)$. This bootstrapping
process leads to more robust detectors.

\subsubsecspcY\subsubsection{Latent variable updates.}

To perform updates on $\ps{M}$ and $\ps{V}$ the learnt detector is
evaluated on the corresponding training images. New values for $m$
and $\pv{v}$ are taken from the detection which gives the highest
score \emph{and} overlaps with the ground-truth bounding box by at
least $70\%$. The new values of latent variables replace the
values of that training example in $\ps{M}$ and $\ps{V}$.

\subsubsecspcY\subsubsection{Relation to other methods.}

The proposed optimization scheme can be interpreted as a Multiple
Instance~(MI)~SVM~\cite{Andrews03} by treating all combinations of
mixture assignments/visibility flags as an (exponentially large)
`bag' of instances for each positive example. Consequently our
method is also closely related to the optimization framework of
the Deformable Part-based Models (DPM) by
Felzenszwalb~\etal~\cite{Felzenszwalb10} and we observe the same
notion of semi-convexity.

\secspcYabove\section{Fusion and Non-Maximum Suppression}\secspcYbelow
\label{sec:nms}
As with conventional sliding window detectors, our proposed method defined so far will tend to give multiple `candidate' detections for a single object due to invariance in the image descriptor. Given a set of such candidates $\ps{D}$
(Fig.~\ref{fig:Overview}~(b)) we require a method for fusing detections into one per ground-truth object. This is essential for evaluation on the PASCAL VOC data~\cite{voc_ijcv}, since the evaluation criterion for a true positive allows only one detection per ground-truth object -- all additional detections are considered false positives. The goal of NMS is to infer a subset $\ps{I} \subseteq \ps{D}$ of final detections, which we refer to as a \emph{scene interpretation}.

\subsubsecspcY\subsubsection{Previous work.}

NMS in sliding window detectors has generally been treated as a
process separate from the rest of the detection framework.
Felzenszwalb~\etal~\cite{Felzenszwalb10} predict bounding boxes
using a linear regressor based on the positions of object parts
and then employ a greedy selection scheme. This scheme iteratively
selects the highest ranking detection and removes all detections
which overlap to a certain degree $\eta$. The overlap measure is
defined as the ratio of the intersection of the bounding boxes to
the area of the lower ranked bounding box. While lacking a clear
theoretical motivation, the main effect is that lower ranked
detections which are fully contained within a higher ranked
detection are removed.

Another popular approach to NMS is to use the mean-shift
technique, treating detections as points in scale-space and
converging on modes of the `distribution' of
detections~\cite{Leibe04,Dalal06}. The method also lacks a clear
theoretical motivation since it is not clear that modes (where the
detector fires in many near-by locations) should be considered
good detections. Compared to the greedy selection strategy of
Felzenszwalb~\etal~\cite{Felzenszwalb10} we have found
experimentally that the method of Dalal~\cite{Dalal06} performs
slightly worse.

Neither scheme acknowledges the possibility of partial visibility
of an object but instead `hallucinates' the full extent of an
object. This is problematic as the assumption of fully visible
objects can cause suppression of partially visible objects. In
addition, neither scheme is guaranteed to converge to a globally
optimal solution: the method of
Felzenszwalb~\etal~\cite{Felzenszwalb10} is greedy by nature while
mean-shift methods are generally non-convex optimization problems.

\subsecspcYabove\subsection{NMS by Detection
Covering}\subsecspcYbelow

The two key components of the NMS scheme proposed here are (i)~a
novel overlap measure, which accounts for the possibility of
partial visibility of an object, and (ii)~an energy function which
offers a way to infer a good scene interpretation alongside an
optimizer which is guaranteed to converge to the global optimum.
The basic idea of our scheme is to find a set of bounding boxes
which cover all detections while taking into account their
inferred visibility.

\begin{figure}
    \figspcYabove
        \centering
        \begin{tabular}{ccc}
            \includegraphics[scale=0.75]{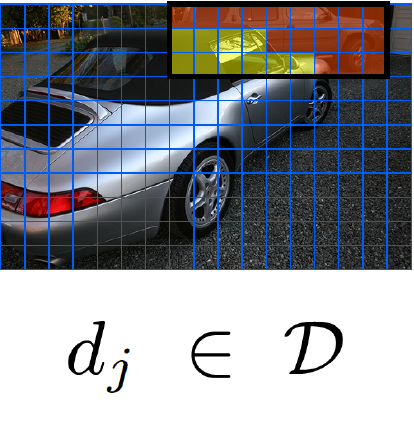}
            &
            \includegraphics[scale=0.75]{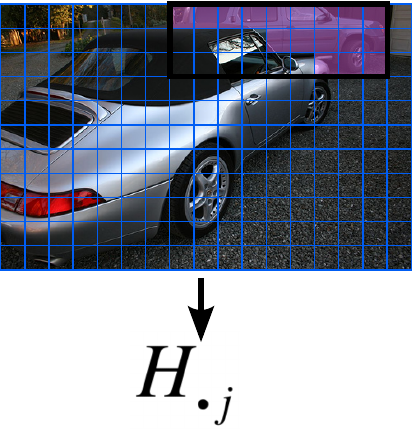}
            &
            \includegraphics[scale=0.75]{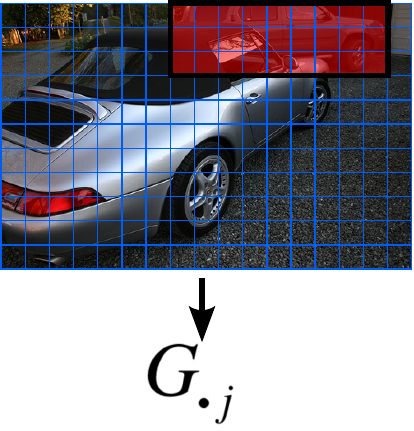}
            \\
       (a) & (b) & (c) \\
        \end{tabular}
    \figspcYcaption
        \caption[Matrices for the non-maximum suppression scheme]{Matrices for the NMS scheme. (a)~a candidate detection $d_j \in \ps{D}$ (black bounding box) with inferred visible (orange) and occluded (yellow) regions; (b)~the visible blocks of $d_j$ are projected to the block grid of the scene. All purple blocks are set to $1$, others to $0$; (c)~The score of $d_j$ is projected to all blocks of the scene that are covered by the \emph{entire} detection. Each orange block is set to the score of $d_j$, others to $0$. Finally, the block grids of (b) and (c) are rasterized into the rows of matrices $\pmm{G}$ and $\pmm{H}$.}
        \label{fig:GH_Overview}
    \figspcYbelow
\end{figure}

The detection framework described in Sec.~\ref{sec:method} outputs
detections $\ps{D}$ at different scales. As a first step we ensure
comparability of these detections by interpolating their
visibility flags to a common scale level using nearest-neighbor
interpolation. This common scale level is represented by a block
grid of size $b_x \times b_y$ and we use the smallest scale level
(a scale factor of $1$). In addition, we normalize the
scores of all detections in $\ps{D}$ by subtracting the threshold
$t$ that was used to obtain the detections, ensuring that the
minimum score will be $0$.

To explain our NMS scheme two matrices $\pmm{H}$ and $\pmm{G}$ are
defined, which relate single detections of $\ps{D}$ to a scene.
Both matrices are of size $b_x b_y  \times \left| \ps{D} \right|$,
where $\left| \ps{D} \right|$ is the number of detections and $b_x
b_y$ is the total number of blocks in a scene -- see
Fig.~\ref{fig:GH_Overview} for a visual explanation. Elements in
$\pmm{H}$ state whether a specific block of the scene is covered by a
detection $d_j \in \ps{D}$. Element $H_{ij}=1$ if the $i$th
feature block \emph{of the scene} is visible in the $j$th
detection, \ie the corresponding value in $\pv{v}^j  \in \ps{V}$
is set to 1 -- see Fig.~\ref{fig:GH_Overview}~(b). The second
matrix $\pmm{G}$ projects the individual detection scores
(Eqn.~\ref{eqn:mixmod}) to the block grid of the scene -- see
Fig.~\ref{fig:GH_Overview}~(c). Elements in $\pmm{G}$ corresponding to
\emph{all} blocks belonging to the $i$th detection -- visible or
not -- are assigned the score of that detection.

\subsecspcYabove\subsection{Overlap Measure}\subsecspcYbelow
Let us define a principled overlap measure between two detections $d_l ,d_k\in \ps{D}$ which considers the number of shared \emph{visible} blocks between both detections:
\begin{equation}
\eqnspcYabove
\vspace{-5pt}
r\left( {d_l ,d_k } \right) = {{\sum\limits_{i = 1}^{b_x b_y } {H_{ik}  \wedge H_{il} } } \mathord{\left/
 {\vphantom {{\sum\limits_{i = 1}^{b_x b_y } {H_{ik}  \wedge H_{il} } } {\sum\limits_{i = 1}^{b_x b_y } {H_{ik}  \vee H_{il} } }}} \right.
 \kern-\nulldelimiterspace} {\sum\limits_{i = 1}^{b_x b_y } {H_{ik}  \vee H_{il} } }}
\label{eqn:ovlmeas}
\eqnspcYbelow
\vspace{2pt}
\end{equation}

This overlap measure is related to existing methods using bounding box intersection as an overlap measure. However, given the availability of inferred partial visibility flags we consider this overlap measure more informative   because it takes the visible portion of the objects into account and does not hallucinate their full extent \cf common overlap measures~\cite{Felzenszwalb10}.

\subsecspcYabove\subsection{Energy Maximization for NMS}\subsecspcYbelow
To infer a good scene interpretation $\ps{I}$ we would like to maximize the accumulated `score' of objects in the scene while compensating for multiple detections per ground-truth object. We therefore express NMS as a constrained energy maximization problem. The final and optimal set of detections $\ps{I}^*$ is inferred by
\begin{equation}
\eqnspcYabove
\begin{aligned}
& \ps{I}^*  = \mathop {\arg \max }\limits_{\ps{I}} F\left( \ps{I} \right)
 \quad \text{where} \quad F \left( \ps{I} \right) = \sum\limits_{i = 1}^{b_x b_y } {\mathop {\max }\limits_{d_j  \in \ps{I}} {G}_{ij} } \\
& \text{subject to: }  \forall d_k ,d_l  \in \ps{I}:r\left( {d_k ,d_l } \right) \le \eta \\
\end{aligned}
\label{eqn:nmsnrg}
\eqnspcYbelow
\end{equation}

This means that for each block on the scene grid the algorithm aims to include the detection that provides the maximum score for that block according to matrix $\pmm{G}$. However, to ensure that the final set of detections is plausible in the context of the scene, a constraint on the maximum amount of overlap between detections $\eta$ is imposed. This can be considered similar to the overlap constraint imposed by the PASCAL VOC challenge~\cite{voc_ijcv}, but accounting for partial visibility.

\subsecspcYabove\subsection{Optimization by Branch-and-Bound}\subsecspcYbelow
\label{chal:occmod:sec:bnb}

The search space of the problem described in Eqn.~\ref{eqn:nmsnrg} consists of all combinations of detections that satisfy the overlap constraint. We explore the search space using a priority search, which starts with an empty scene interpretation having zero energy. We then iteratively pick the highest ranking interpretation and add one additional detection (satisfying the overlap constraint), recalculate the energy and add it to the queue of already computed scene interpretations.

If $\left| \ps{D} \right|$ is relatively small, a brute-force search is possible. However, in the case that the number of detections is large, a more efficient search procedure is necessary. We propose a novel branch-and-bound procedure by observing that for any given scene interpretation $\ps{S}$ the following upper bound for the remaining detections still satisfying the overlap constraint of Eqn.~\ref{eqn:nmsnrg} can be computed:
\begin{equation*}
\eqnspcYabove
\hat F \left( \ps{S} \right) = \sum\limits_{i = 1}^{b_x b_y } {\mathop {\max }\limits_{d_j  \in \ps{S} \cup \ps{C}} G_{ij} }  \ge F \left( \ps{S} \right) \quad \text{where} \quad \ps{C} = \left\{ {\left. d \right|d \in \ps{D} \wedge \forall s \in \ps{S}:r\left( {d,s} \right) \le \eta } \right\}
\eqnspcYbelow
\end{equation*}
\vspace{2pt} where $\ps{C}$ is the set of detections that
satisfies the constraint of Eqn.~\ref{eqn:nmsnrg} given the
already selected detections $\ps{S}$. This upper bound answers the
question of what energy could possibly be achieved if we continue
adding detections to this scene interpretation. If it is smaller
than the current best energy, the scene interpretation does not
need to be explored further. In contrast to other NMS
schemes~\cite{Felzenszwalb10,Dalal06} the proposed scheme is
guaranteed to converge to a global maximum. To further reduce the
optimization time for a large set of detections $\ps{D}$ we prune
very weak detections which score less than some confidence
threshold $t=-1$.

\secspcYabove\section{Empirical Results}\secspcYbelow
\label{sec:results}
This section reports experimental results -- we first discuss
implementation details, then report quantitative and qualitative
results on PASCAL VOC 2010~\cite{voc_ijcv}.

\subsubsecspcY\subsubsection{Window descriptor.}
We use the HOG descriptor~\cite{Dalal05}, which computes local histograms of gradient orientation in a set of square `cells' laid out on a regular grid. Adjacent cells are then aggregated and normalized to give `blocks' with greater invariance to local lighting and spatial deformation. As discussed earlier, the block structure of these features provides a suitable basis for building detectors allowing for partial visibility by inferring visibility at the block level.

\subsubsecspcY\subsubsection{Parameter estimation.}
Cross-validation is used to set parameters $\left\{
{\lambda,\alpha ,\eta } \right\}$. To set the regularization
parameter $\lambda$ we learn detectors without partial visibility
modelling and consider possible values in the range of $10^{ - 1}$
to $10^{2}$. We adopt the value which gives the highest validation
AP at $\eta=0.5$ using the greedy NMS scheme of
Felzenszwalb~\etal~\cite{Felzenszwalb10}. We then fix this value
of $\lambda$ and use cross-validation to determine the best
overlap threshold $\eta$ for the greedy NMS scheme in the range of
possible values $0.2$ to $1.0$. The best set of parameters $\left(
{\lambda ^* ,\eta ^* } \right)$ forms the baseline detector. To
determine the strength of the contiguity term $\alpha$ we fix
$\lambda$ to $\lambda ^*$ and perform cross-validation in the
range of $10^{ - 2}$ to $10^{ - 1}$. Again, we pick the value
${\alpha ^* }$ which yields best AP using the greedy NMS scheme.
Finally, the overlap threshold $\eta$ for the NMS scheme presented
in Sec.~\ref{sec:nms} is determined by cross-validation
considering possible values in the range of $0.2$ to $1.0$.

\subsubsecspcY\subsubsection{Truncated Objects.}
To detect truncated objects we pad the feature representation of training and test images on each side with $50\%$ of the window size at each scale. These padded areas have their visibility flags $\pv{v}$ set to $0$, \ie we consider these areas not visible and do not infer them~\cite{voc-release4}.

\subsubsecspcY\subsubsection{Training~\&~Test Protocol.}
Initial mixture assignments $\ps{M}$ are computed by performing $k$-means clustering with $d$ clusters on the bounding box ratio. $d$ is the number of mixture components. Initially only fully visible object instances are used, \ie all visibility flags are set to $1$.

Detectors were learnt using $6$ updates of the latent variables with each utilizing $20$ bootstrapping rounds to learn the model variables $\ps{O}$ using $200$ randomly selected training images. We stop bootstrapping if the number of false positives drops below $200$ for each mixture component. Throughout training and testing an image pyramid with a scale factor of $1.2$ is used. The stride of the sliding window detector is set to 6 pixels.

\subsubsecspcY\subsubsection{Average Bounding Box.}
We predict an average bounding box per mixture component, which is chosen in such a way that it maximizes overlap with all ground-truth bounding boxes assigned to a mixture component. If the visibility flags indicate that the object is smaller than the average bounding box, we contract the bounding box around the visibility flags.

\subsecspcYabove\subsection{Quantitative Results}\subsecspcYbelow
\label{sec:quantres}

For performance evaluation we use the PASCAL VOC 2010~\cite{voc_ijcv} datasets and methodology -- precision/recall curve, reporting Average Precision (AP), with bounding box overlap of 50\%. We
compare different subsets of our method.

\subsubsecspcY\subsubsection{Nomenclature.} The baseline (short:
\emph{BL}) is a detector with $2$ mixture components but no
visibility modelling, using the greedy NMS scheme of
Felzenszwalb~\etal~\cite{Felzenszwalb10}. \emph{VIS} additionally
models partial visibility and also uses the greedy NMS scheme.
\emph{VIS+NMS} additionally employs the proposed NMS scheme.

\subsubsecspcY\subsubsection{Partial Visibility Modelling.}
\begin{figure}
    \figspcYabove
        \centering
        \begin{tabular}{cccc}
        \hspace{-4pt} \includegraphics[scale=0.23]{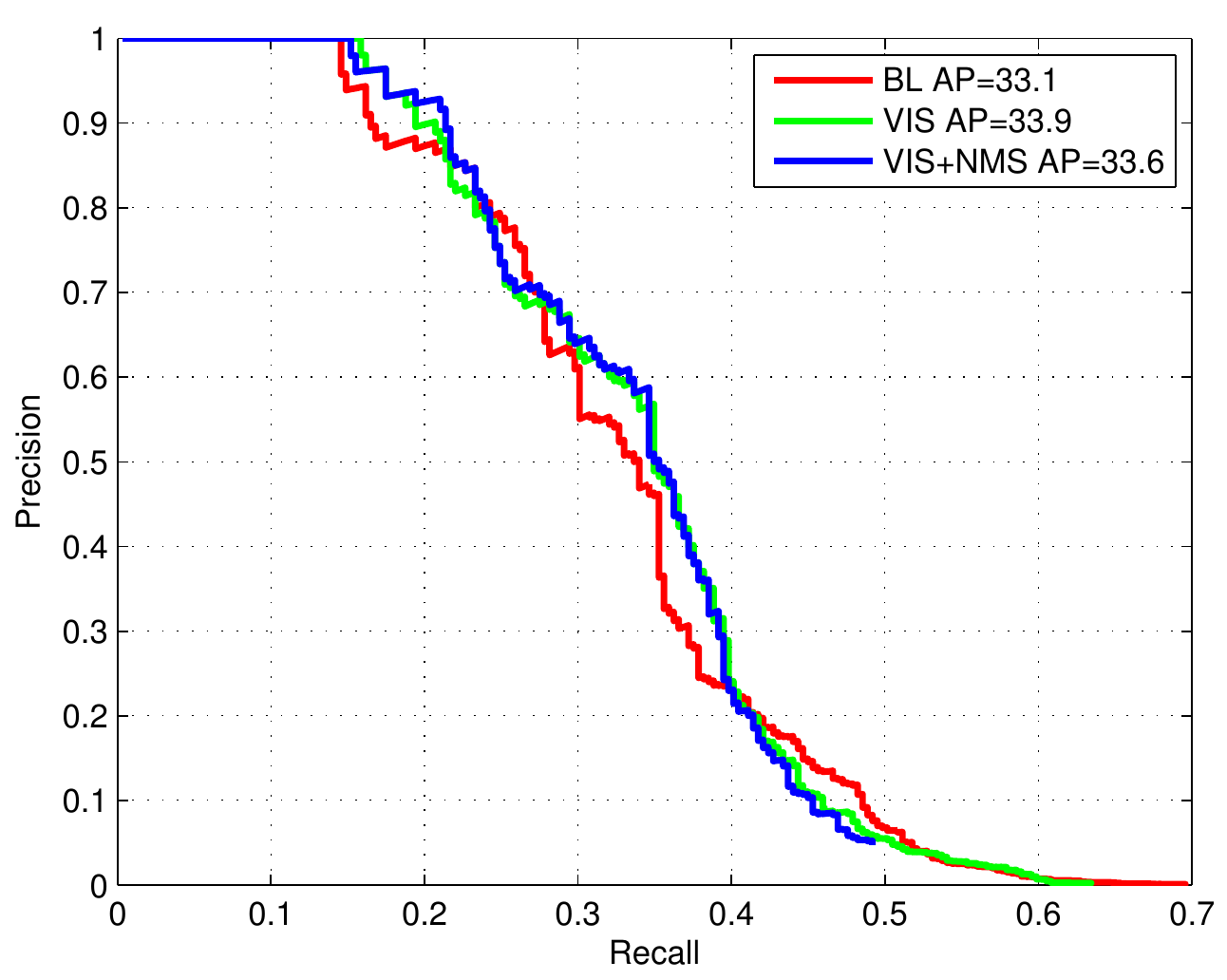}
            &
           \hspace{-5pt}  \includegraphics[scale=0.23]{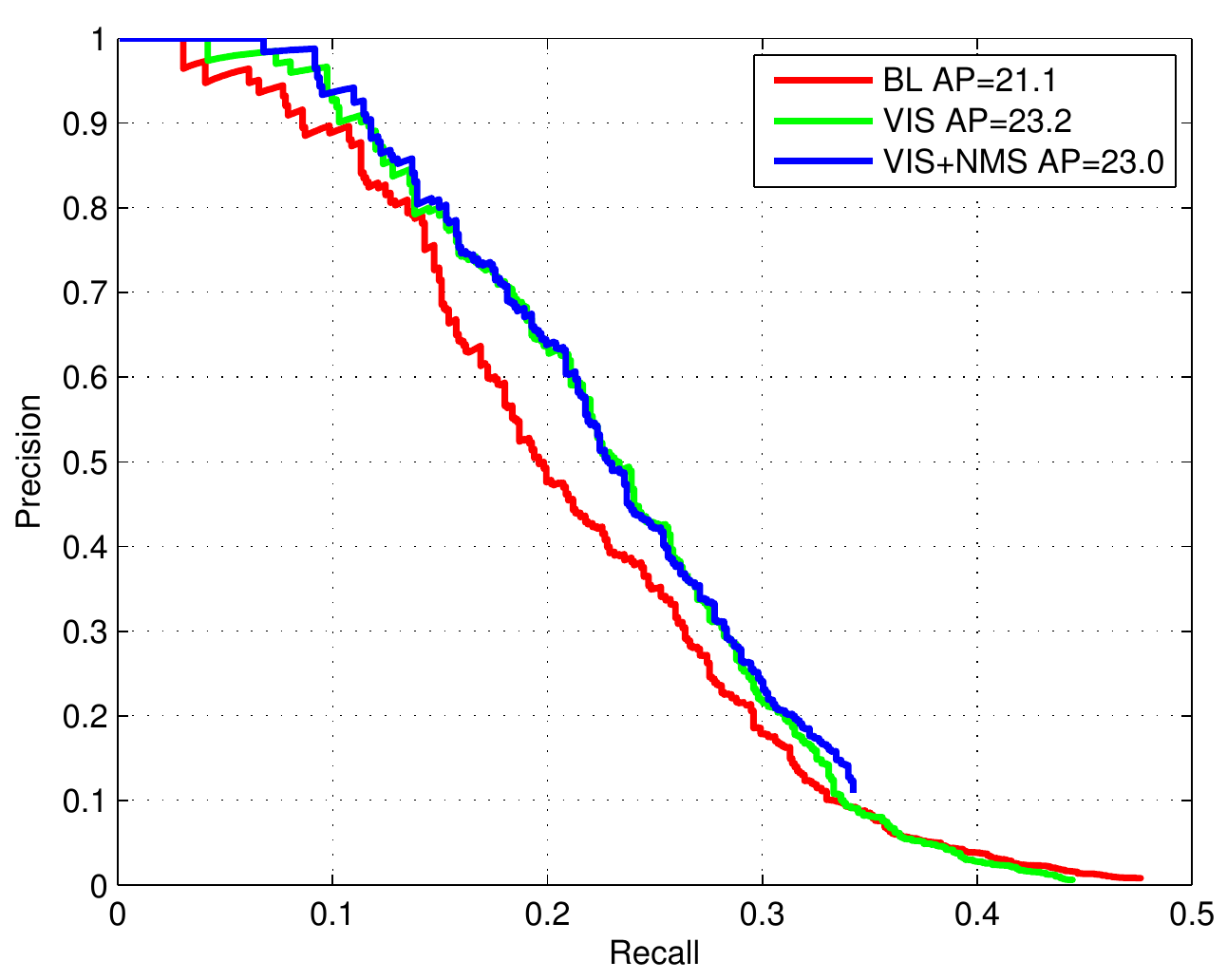}
            &
            \hspace{-5pt}  \includegraphics[scale=0.23]{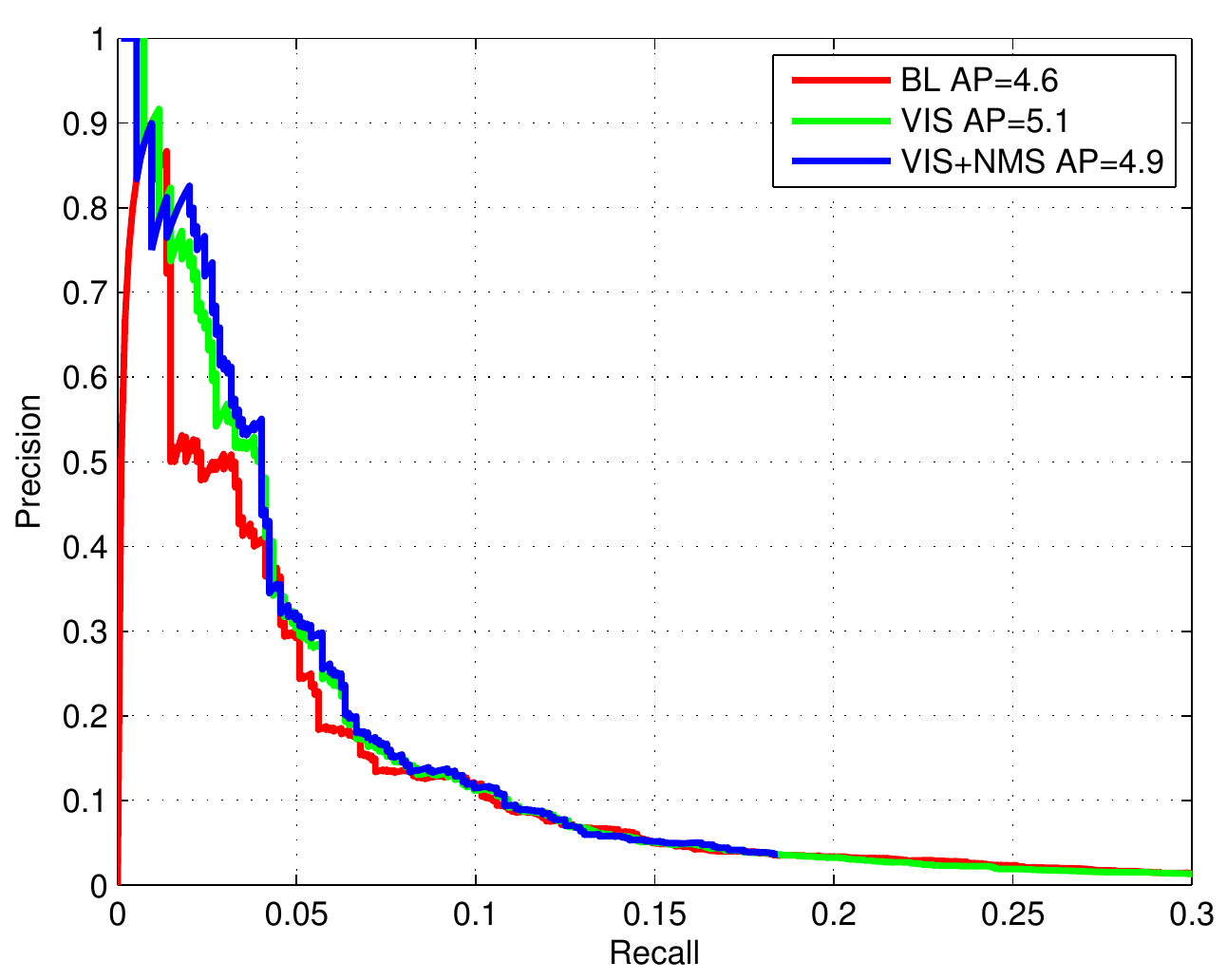}
            &
            \hspace{-5pt} \includegraphics[scale=0.23]{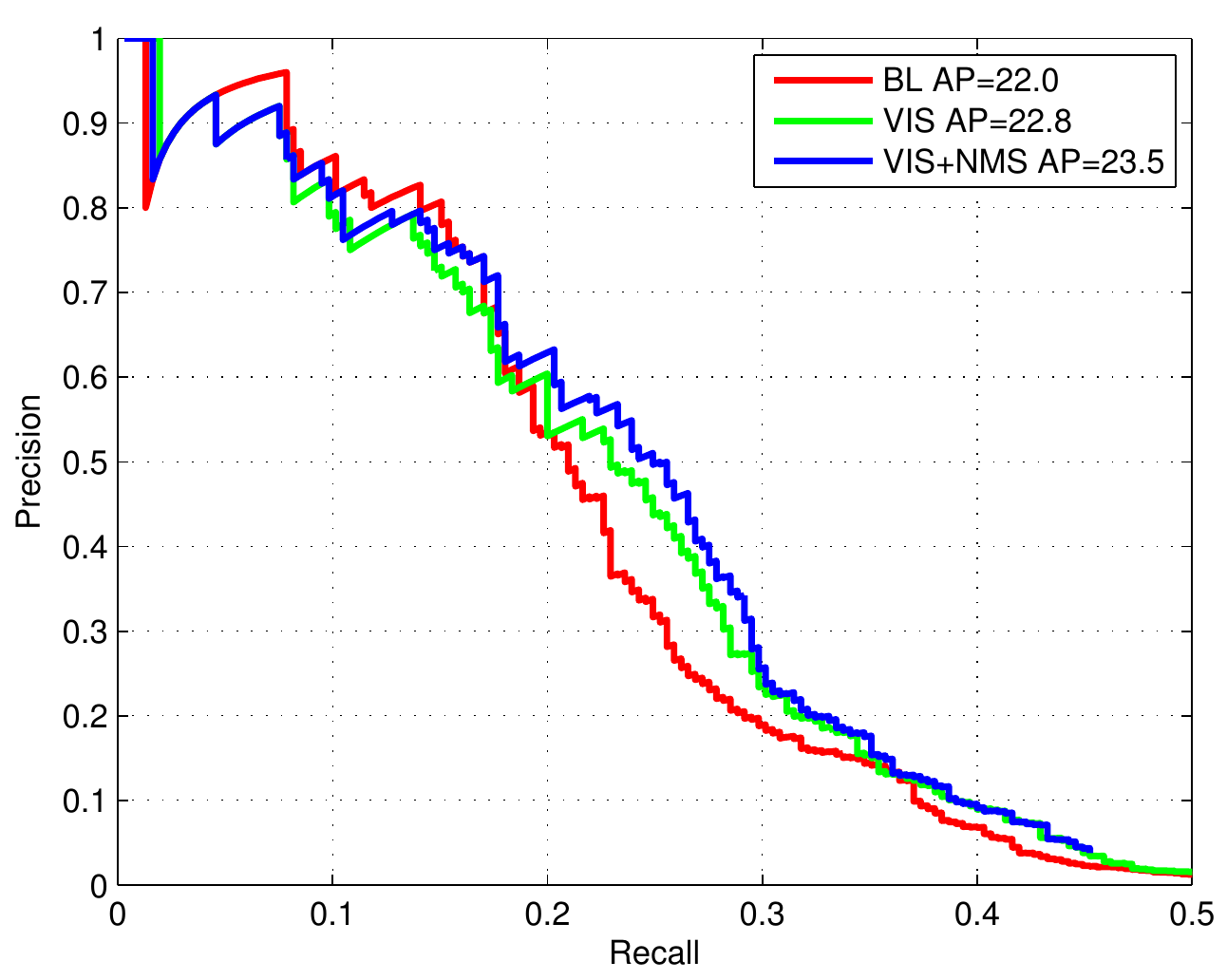}
            \\
        (a)~bicycle & (b)~car & (c)~chair & (d)~motorbike
        \end{tabular}
    \figspcYcaption
        \caption[Precision/recall curves for selected PASCAL VOC 2010 classes]{
        Precision/recall curves for selected PASCAL VOC 2010 classes, comparing \emph{BL} (red curves) to \emph{VIS} (green) and \emph{VIS+NMS} (blue). Modelling of partial visibility gives an increased recall at comparable precision levels for all shown classes.}
        \label{fig:ResultPlotsVOC}
    \figspcYbelow
\end{figure}

Table~\ref{tab:2010res} shows the AP results comparing \emph{BL}
to \emph{VIS}. Partial visibility modelling improves AP for $16$
of $20$ classes while mean AP improves from $12.2\%$ to $12.9\%$.
AP improves for example for the `car' class from $22.3\%$ to
$23.5\%$ when comparing \emph{BL} to \emph{VIS} (a relative
improvement of $5.4\%$) or for the `horse' class, which improves
from $13.4\%$ to $14.9\%$ ($11.2\%$ relative).
Fig.~\ref{fig:ResultPlotsVOC} furthermore presents
precision/recall curves for four classes of the VOC 2010 dataset,
establishing that for these (and most other) classes partial
visibility modelling gives an increased recall at almost all
precision levels.

\input{res_occmod_table}

\subsubsecspcY\subsubsection{Globally Optimal NMS Scheme.} The
proposed NMS scheme gives comparable precision/recall curves to
the greedy NMS scheme~\cite{Felzenszwalb10} -- see blue curves in
Fig.~\ref{fig:ResultPlotsVOC}. We also evaluate the \emph{VIS}
experiment at $t=-1$ and compare the results to \emph{VIS+NMS} in
Table~\ref{tab:2010res} to allow for a fair comparison as
\emph{VIS+NMS} uses a higher threshold for efficiency reasons. The
results can be interpreted as how the two different NMS schemes
perform when presented with the \emph{same} set of detections
$\ps{D}$ per image. Table~\ref{tab:2010res} shows that overall
there is no clear improvement as mean AP decreases slightly from
$12.5\%$ to $12.3\%$. AP improves for $6$ object classes and
remains the same for another $6$ object classes. We believe the
new NMS scheme will prove beneficial with improved modelling of
partial visibility.

\subsecspcYabove\subsection{Qualitative Results}\subsecspcYbelow
\label{sec:qualres}

In the following we demonstrate the applicability of partial visibility modelling to object detection by example detections as qualitative
results.

\subsubsecspcY\subsubsection{Partial Visibility Modelling.}
\begin{figure}
    \figspcYabove
        \centering
        \begin{tabular}{cccc}
        \includegraphics[scale=0.14]{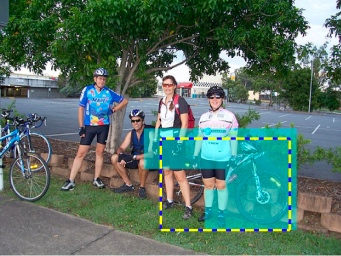}   \hspace{-5pt}
            &
            \includegraphics[scale=0.14]{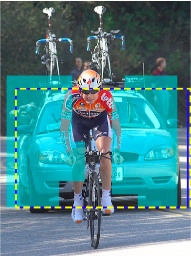}  \hspace{-5pt}
            &
            \includegraphics[scale=0.14]{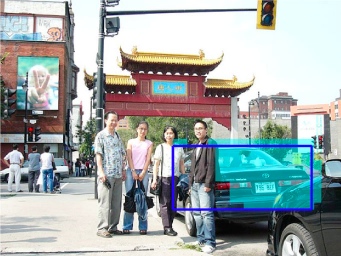} \hspace{-5pt}
            &
            \includegraphics[scale=0.14]{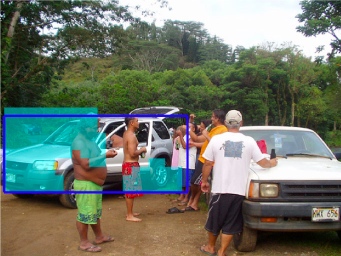}
            \\
            (a)~bicycle & (b)~car & (c)~car & (d)~car
            \\
            \includegraphics[scale=0.14]{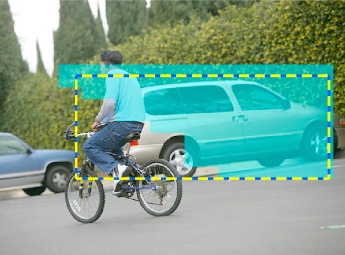} \hspace{-5pt}
            &
             \includegraphics[scale=0.14]{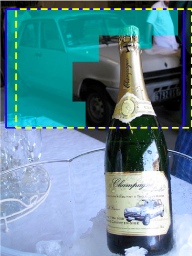} \hspace{-5pt}
            &
            \includegraphics[scale=0.14]{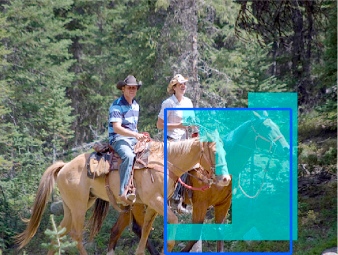} \hspace{-5pt}
            &
            \includegraphics[scale=0.14]{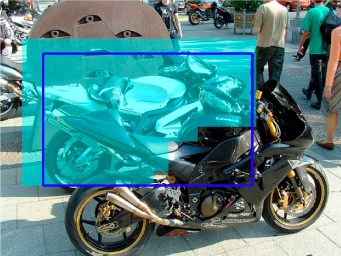}
            \\
            (e)~car & (f)~car & (g)~horse & (h)~motorbike
        \end{tabular}
    \figspcYcaption
        \caption[Example detections for the baseline and partial
        visibility modelling]{Example detections for the baseline
        and partial visibility modelling. We show detections from
        the \emph{BL} and \emph{VIS} experiments. Where detected,
        baseline detections are shown as bounding boxes with
        dashed yellow lines. Our method is able to infer the
        visible extent of an object (cyan regions) \cf the
        baseline, which does not have this capability. To avoid
        visual clutter only one detection per image is shown. }
        \label{fig:ExampleDetections}
    \figspcYbelow
\end{figure}

Fig.~\ref{fig:ExampleDetections} shows example detections for four
object classes. Compared to \emph{BL} (dashed yellow bounding
boxes), \emph{VIS} (solid blue bounding boxes) is able to infer
partial visibility -- active visibility flags are shown in cyan.
Our method detects a wide range of occlusions \eg in (b) the
person occludes the middle part of the car and these blocks are
correctly inferred as not visible; in (f) the bottle occludes
large parts of the car.

\subsubsecspcY\subsubsection*{Globally Optimal NMS Scheme.}
\begin{figure}
    \figspcYabove
        \centering
        \begin{tabular}{cccc}
            \includegraphics[scale=0.121]{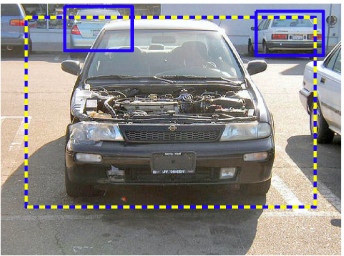}   \hspace{-5pt}
            &
            \includegraphics[scale=0.121]{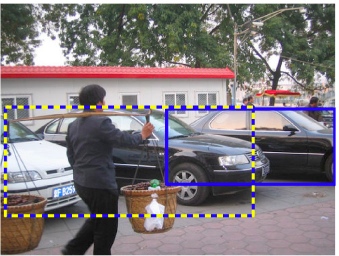}  \hspace{-5pt}
            &
           \includegraphics[scale=0.121]{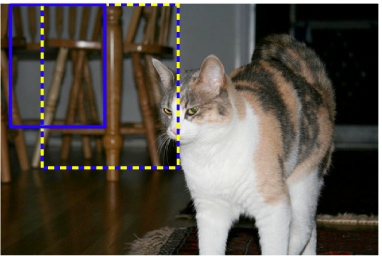}  \hspace{-5pt}
        &
        \includegraphics[scale=0.121]{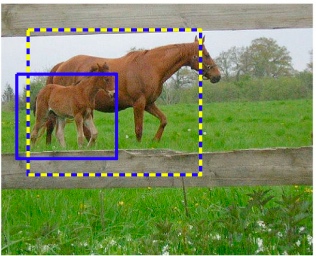}
            \\
            (a)~car & (b)~car  & (c)~chair & (d)~horse
             \\
            \includegraphics[scale=0.121]{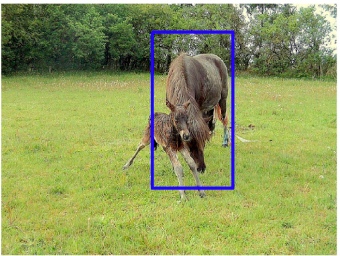} \hspace{-5pt}
            &
            \includegraphics[scale=0.121]{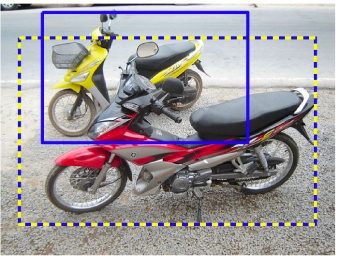} \hspace{-5pt}
        &
        \includegraphics[scale=0.121]{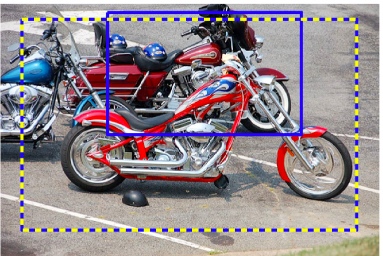} \hspace{-5pt}
        &
            \includegraphics[scale=0.121]{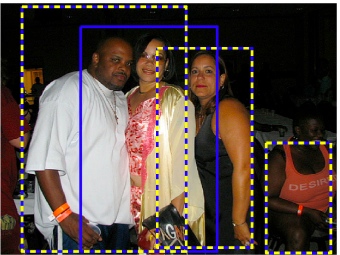}
            \\
            (e)~horse & (f)~motorbike & (g)~motorbike & (h)~person
        \end{tabular}
    \figspcYcaption
        \caption[Additional detections of the proposed NMS
        scheme]{Additional detections of the proposed NMS scheme
        (\emph{VIS+NMS}). Where detected, \emph{VIS}$_{t=-1}$
        detections are drawn as yellow bounding boxes with dashed
        lines. The new NMS scheme yields additional occluded
        detections (shown in blue).}
        \label{fig:ExampleNMS}
    \figspcYbelow
\end{figure}

In Fig.~\ref{fig:ExampleNMS} we compare detections from the \emph{VIS}$_{t=-1}$ to the \emph{VIS+NMS} detector. The proposed NMS scheme yields additional occluded detections because it employs a principled overlap measure (Eqn.~\ref{eqn:ovlmeas}). The new NMS scheme is also able to extract entirely new non-occluded detections in comparison to the greedy scheme \eg (h) due to the proposed block-based energy function (Eqn.~\ref{eqn:nmsnrg}).

In summary, the new NMS scheme is able to extract additional
detections at the cost of a slightly lower precision rate, \ie it
also extracts more false-positives on average. The positive
aspects of the proposed NMS scheme remain: (i)~it is more
principled than the greedy NMS scheme~\cite{Felzenszwalb10} in
that it takes partial visibility into account; (ii)~provides a
globally optimal solution and (iii)~it is general, \ie we believe
it can be improved with better modelling of partial visibility.

\secspcYabove\section{Conclusions}\secspcYbelow
\label{sec:conclusions}

We have proposed a method for detection of partially visible
objects in a sliding-window framework. Key contributions of the
method are the treatment of partial visibility as a latent
variable, a NMS scheme which takes into account partial
visibility, and the ability to correctly predict the visible
extent of an object (Fig.~\ref{fig:ExampleDetections}) \cf
state-of-the-art approaches~\cite{Felzenszwalb10,Vedaldi09}. In
future work we plan to combine our scheme with a stronger baseline
such as part-based models~\cite{Felzenszwalb10} and make use of
more powerful window descriptors~\cite{Vedaldi09,Ott09,Wang09}.

\vspace{-12pt}
\bibliographystyle{splncs}
\bibliography{eccv2012}

\end{document}

%% file: res_occmod_table.tex
\begin{table}
\vspace{-5pt}
 \begin{tabular}{| >{\centering\arraybackslash}m{45pt} | 
 									 >{\centering\arraybackslash}m{26pt} |
 									 >{\centering\arraybackslash}m{26pt} |
 									 >{\centering\arraybackslash}m{26pt} |
 									 >{\centering\arraybackslash}m{26pt} |
 									 >{\centering\arraybackslash}m{26pt} |
 									 >{\centering\arraybackslash}m{26pt} |
 									 >{\centering\arraybackslash}m{26pt} |
									 >{\centering\arraybackslash}m{26pt} |
									 >{\centering\arraybackslash}m{26pt} |
									 >{\centering\arraybackslash}m{26pt} |
 								}

	\hline
   AP 
			& 			aero & 		bike & 		bird & 		boat & 		bottle & 		bus & 		car & 		cat & 		chair &		cow  \\
   \hline
   \emph{BL}	& 			25.7 & 	    	\pv{28.8} & 	0.2 & 		2.4 & 		0.9 & 		30.5 & 		22.3 & 		2.5 & 		5.1&			3.9  \\
   \emph{VIS}	& 	 		\pv{28.0} &	27.9 & 		0.2 & 		\pv{3.4} & 	\pv{1.3} &		32.3 &		23.5	& 		\pv{3.9} &		\pv{5.9} &		\pv{5.4} \\
   \emph{VIS}$_{t=-1}$	& 	27.7 & 	    	27.7 & 		0.1 & 		3.2 & 		1.0 & 		31.9 & 		22.9 &		3.0 & 		5.6 & 		5.2 \\
   \emph{VIS+NMS}	& 	 	26.3 &	 	28.1 & 		0.1 & 		3.2 & 		1.0 &	 	\pv{32.5} &	\pv{23.7} &	3.0 &	 	5.8 &	 	5.3 \\
   \hline \hline
   AP
			& 		table & 		dog & 		horse & 		pers &		mbike & 		plant & 		sheep & 		sofa & 		train & 		tv \\
   \hline
   \emph{BL}	& 		5.2 & 		3.7 & 		13.4 & 		\pv{20.5} & 	25.6 & 		3.4 & 		7.1 & 		3.6 & 		13.2 & 		26.3 \\
   \emph{VIS}	& 		\pv{5.3} &		\pv{4.6} &		\pv{14.9} &	17.6 &		\pv{27.3} &	\pv{3.7} &		\pv{8.1} &		\pv{3.8} &		13.2 & 		\pv{27.3} \\
   \emph{VIS}$_{t=-1}$& 	4.9 & 		3.9 & 	 	14.4 & 	 	16.2 &		27.1 &		3.5 &		7.8 &		3.3 &		12.8 & 		26.9 \\
   \emph{VIS+NMS}	& 	4.3 &		3.9 &		13.7 &		15.8  &		27.2 &		3.5 &		7.5 &		3.2 &		12.2 & 		25.9\\
   \hline
  \end{tabular}
  \vspace{2pt}
  \caption[AP results for PASCAL VOC 2010 (partial visibility modeling) I]{PASCAL VOC 2010 results ({\texttt{\small{comp3}}}, {\texttt{\small{test}}}-set). We compare AP for models with and without visibility modeling (\emph{VIS} vs. \emph{BL}) using the greedy NMS scheme~\cite{Felzenszwalb10}. We also compare the greedy NMS scheme (\emph{VIS}$_{t=-1}$) to the new NMS scheme (\emph{VIS+NMS}) at a detector threshold of $t=-1$. Best AP is shown in bold. See text for discussion.}
  \label{tab:2010res}
\vspace{-8pt}
\end{table}